\documentclass{article}

\usepackage{microtype}
\usepackage{graphicx}
\usepackage{subfigure}
\usepackage{booktabs} 

\usepackage{balance}
\usepackage{xurl}

\usepackage{hyperref}

\usepackage[accepted]{icml2025}

\usepackage{amsmath}
\usepackage{amssymb}
\usepackage{mathtools}
\usepackage{amsthm}

\usepackage[capitalize,noabbrev,nameinlink]{cleveref}

\theoremstyle{plain}

\theoremstyle{definition}

\theoremstyle{remark}

\usepackage[textsize=tiny]{todonotes}

\usepackage{enumitem}
\usepackage{cleveref}
\usepackage{tikz}
\usepackage{listings}
\usepackage{xcolor}
\usepackage{pifont}
\usepackage{multirow}

\usepackage[table]{xcolor}
\usepackage{booktabs}
\usepackage{transparent}

\definecolor{red}{RGB}{180,12,13}
\definecolor{blue}{RGB}{51,138,158}
\definecolor{green}{RGB}{51,188,138}

\newcommand{\header}[1]{\vspace{1mm}\noindent\textbf{#1}.}
\newcommand{\headernp}[1]{\vspace{1mm}\noindent\textbf{#1}}
\newcommand{\headerl}[1]{\vspace{1mm}\noindent\textit{#1}.}
\newcommand{\method}[1]{#1}

\newcommand*\circled[1]{\protect\tikz[baseline=(char.base)]{\protect\node[shape=circle,fill=red,color=red,draw,inner sep=0.6pt] (char) {\textcolor{white}{\footnotesize \textbf{#1}}};}}

\newcommand{\acc}[1]{#1}

\newcommand{\ri}[1]{{\raisebox{0.3em}{#1}}}

\lstdefinestyle{promtStyle}{
    basicstyle=\ttfamily\footnotesize,
    backgroundcolor=\color{gray!10},
    breaklines=true,
    breakindent=0pt,
    postbreak={},
    showstringspaces=false,
    captionpos=b,
    keywordstyle=\color{blue}\bfseries,
}

\newcommand{\xmark}{\ding{55}}  
\usepackage{numprint}
\newcommand{\num}[1]{\numprint{#1}}
\npstyleenglish

\icmltitlerunning{Towards Cross-Modal Error Detection with Tables and Images}

\begin{document}

\twocolumn[
\icmltitle{Towards Cross-Modal Error Detection with Tables and Images}



\icmlsetsymbol{equal}{*}

\begin{icmlauthorlist}
\icmlauthor{Olga Ovcharenko}{bifold}
\icmlauthor{Sebastian Schelter}{bifold}
\end{icmlauthorlist}

\icmlaffiliation{bifold}{BIFOLD \& TU Berlin}

\icmlcorrespondingauthor{Olga Ovcharenko}{ovcharenko@tu-berlin.de}
\icmlcorrespondingauthor{Sebastian Schelter}{schelter@tu-berlin.de}


\vskip 0.3in
]



\printAffiliationsAndNotice{}  


\begin{abstract}
 Ensuring data quality at scale remains a persistent challenge for large organizations. Despite recent advances, maintaining accurate and consistent data is still complex, especially when dealing with multiple data modalities. Traditional error detection and correction methods tend to focus on a single modality, typically a table, and often miss cross-modal errors that are common in domains like e-Commerce and healthcare, where image, tabular, and text data co-exist.
To address this gap, we take an initial step towards cross-modal error detection in tabular data, by benchmarking several methods. Our evaluation spans four datasets and five baseline approaches. Among them, Cleanlab, a label error detection framework, and DataScope, a data valuation method, perform the best when paired with a strong AutoML framework, achieving the highest F1 scores. Our findings indicate that current methods remain limited, particularly when applied to heavy-tailed real-world data, motivating further research in this area.
\end{abstract}


\section{Introduction}
\label{sec:introduction}

Maintaining high-quality data is a challenging task for large organizations and enterprises~\cite{stonebraker2018data,oala2023dmlr,abedjan2016detecting,Singh25_DataVinci}, especially when a high level of automation is required~\cite{mahdavi2019raha, SiddiqiKB23_SAGA,Yan24_GIDCL}. Erroneous data can lead to devastating economic, societal, and scientific consequences, especially in combination with machine learning (ML) methods~\cite{sambasivan2021everyone,mcgregor2021preventing,holstein2019improving,northcutt2021pervasive,birhane2021multimodal}. Consequently, significant resources have been invested into automating data error detection and correction processes, e.g., via data validation systems such as TensorFlow Data Validation (TFDV)~\cite{polyzotis2019data} (deployed at Google) or Amazon Deequ~\cite{schelter2018automating} which is used in several AWS services~\cite{nigenda2022amazon}.

\begin{figure}
\centering
\includegraphics[width=\columnwidth]{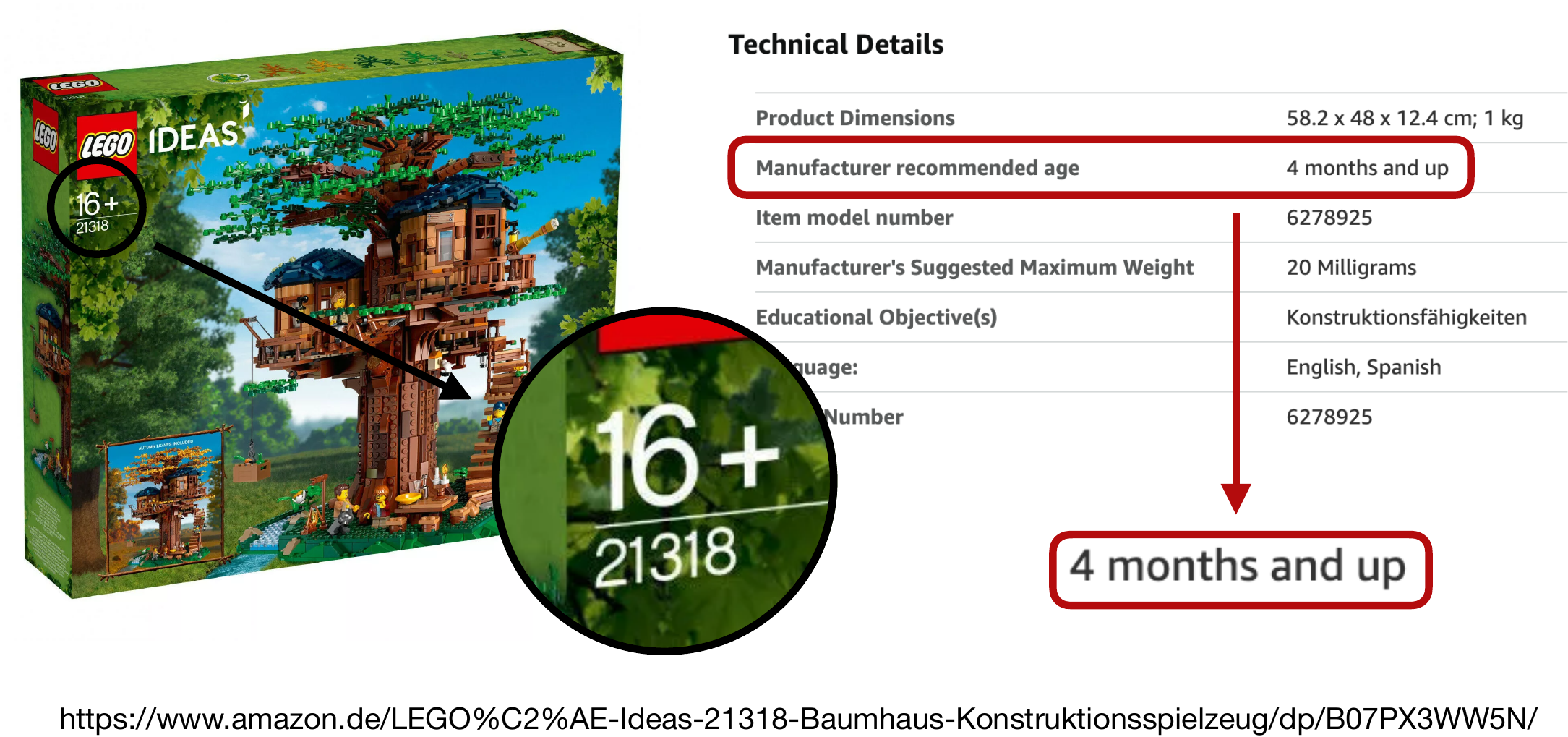}
\vspace{-0.3cm}

\caption{A real-world cross-modal error from an e-Commerce catalog where a toy with a choking hazard is advertised to children of a wrong age group. The tabular product data erroneously states that the LEGO toy is suitable for four month old babies, even though the product image indicates that it is meant for teenagers with 16+ years of age (accessed May 18, 2025).}
\vspace{-0.1cm}

\label{fig:teaser}
\end{figure}

\header{Cross-modal error detection} However, existing data validation systems often focus on a singular modality only (e.g., relational data) and do not cover scenarios where errors may occur across different modalities. Such cross-modal data errors entail inconsistent information across modalities, while each modality alone appears correct. For instance, cross-modal errors occur in online platforms for e-Commerce, traveling, or real estate, as well as self-driving vehicles and electronic health records, where multi-modal data combines tabular data, text, images, and videos. 

\begin{figure*}[!t]
\centering
\includegraphics[width=\textwidth]{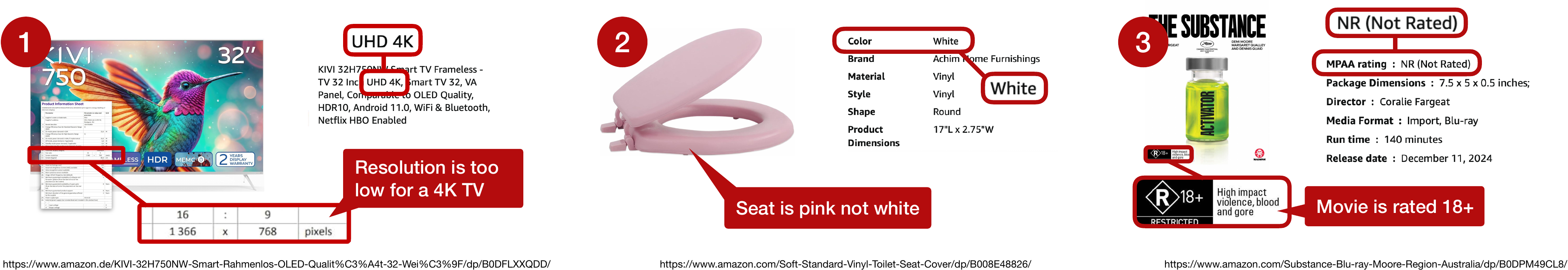}
\vspace{-0.6cm}
\caption{Additional real-world examples of cross-modal errors in Amazon's tabular product metadata, which are obvious from the corresponding product images: \circled{1} A television is misleadingly labeled as 4K ready when the product image shows that its resolution is too low for 4K; \circled{2} A pink toilet seat is listed as having color ``white''; \circled{3} A movie is marked as not rated, even though its cover clearly indicates that it is meant for adult audiences only; (product pages accessed on May 10, 2025).}

\vspace{-0.3cm}

\label{fig:examples}
\end{figure*}

\headerl{Real-world example} To showcase a cross-modal data error from the e-Commerce domain, we refer to~\Cref{fig:teaser}, where a LEGO toy on Amazon is assigned to children of the wrong age group. The tabular product data erroneously states that the LEGO toy is suitable for four-month-old babies, even though the product image indicates that it is meant for 16+ years old teenagers. This data error is potentially dangerous since a toy with such small parts can pose a choking hazard to young children. We point to more cross-modal data errors from Amazon in \Cref{fig:examples}, and would like to highlight that it took us only a couple of minutes to manually find such cases, even though Amazon deploys highly sophisticated methods to manage product attributes~\cite{LinHFZLXD21_PAM}.

\header{Practical challenges} To expand our understanding of this area, we interviewed the content quality team of a large e-Commerce platform. Their product catalog contains tens of millions of products with thousands of distinct attributes. The majority of products originate from several thousand external sellers who provide images and tabular metadata of varying quality and quantity. Moreover, the data from external sellers is combined with additional data sourced from third-party data providers~\cite{YangWYKSSEK22, hou2024bridging}. 
The company's quality team designs custom large language model (LLM)-based solutions for cross-modal error detection and correction of a few selected attributes. However, the high customization requirements and training/inference cost of the attribute-tuned models make their current solution expensive and difficult to scale to more specialized attributes and products.

\headernp{Are existing error detection approaches sufficient?} The detection and repair of errors across diverse data modalities represent an emerging direction in data-centric AI research, and general techniques that work across multiple domains are required. Several existing approaches can be applied to multi-modal error detection, yet it is an open question whether they provide sufficient performance since the majority of them have not been explicitly designed for the cross-modal setting. 

Single-modal tabular error detection approaches~\cite{mahdavi2019raha, holodetect, krishnan2016activeclean} identify inconsistencies between columns in a table but have no direct means to incorporate cross-modality information.
Label error detection methods~\cite{northcutt2021confident, northcutt2021pervasive} focus primarily on label information in predictive settings and would be costly to apply in our scenario since training a specialized model per table attribute is required. LLM-based approaches to error detection~\cite{Hang24_FineMatch, Singh25_DataVinci} show promising results for text-image pairs, but it is unclear how effective they are in handling tabular data combined with visual information. Additionally, LLM approaches suffer from high computational costs and typically require multiple expensive calls to identify erroneous data, which limits their practical application in large-scale data cleaning scenarios. We discuss the mentioned approaches in more detail in \Cref{sec:related_work}.

\header{Overview and contributions} The goal of this paper is to introduce the problem of cross-modal error detection in tabular data and to motivate its high practical importance. Our detailed contributions include:

\begin{itemize}[noitemsep,leftmargin=*]
    \item We motivate and introduce the problem of cross-modal error detection in tabular data~(Sections~\ref{sec:introduction}~\&~\ref{sec:problem}).
    \item We design a preliminary benchmark with four e-Commerce datasets to evaluate five state-of-the-art error detection methods. According to our findings, approaches designed for label error detection, such as Cleanlab~\cite{northcutt2021confident} and DataScope~\cite{karlavs2023data} (combined with the AutoML framework AutoGluon~\cite{tang2024autogluon}), perform the best. Crucially, in the majority of cases, leveraging both tabular and image data is key to uncovering cross-modal errors. Nonetheless, current methods remain limited, particularly when applied to heavy-tailed real-world data.~(\Cref{sec:experiments}).
    
    \item The benchmark containing our data and code is available at: {\small{\url{https://github.com/OlgaOvcharenko/find_errors}}}
\end{itemize}

\section{Problem Statement}
\label{sec:problem}

In this section, we formalize the problem of cross-modal error detection in tables and review related work.

\subsection{Cross-Modal Error Detection in Tables}
\label{sec:problem-definition}

Given aligned multi-modal data $(D, I)$, where $D$ is a relational table and $I$ is a set of corresponding images, the goal is to identify erroneous entries in the relational data $D$. Following \citeauthor{holodetect} (\citeyear{holodetect}), we denote $A = {A_1, A_2, \dots ,A_N }$
the attributes of $D$. We consider $D$ to be a set of tuples, where each tuple $t \in D$ consists of cells $C_t = \{t[A_1],t[A_2], \dots ,t[A_N]\}$. Moreover, $t[A_k]$ denotes the
value of attribute $A_k$ for tuple $t$, and the corresponding image for $t$ is $i_t$. We assume that errors appear due to inaccurate cell assignments in $D$. 
More formally, for a cell $c \in C_t$ we denote by $v_c^*$ its unknown true value and by $v_c$ its observed value. 
We define an erroneous tuple $t \in D$ as a tuple with at least one cell  $c \in C_t$ where $v_c \neq v_c^*$. 

We define cross-modal error detection as deciding whether a tuple $t \in D$ is erroneous, based on its tabular data $C_t$ and corresponding image $i_t$. Importantly, we assume a setup where no labeled examples of erroneous records are available as training data, akin to novelty detection~\cite{pimentel2014review}.

\subsection{Related Work}
\label{sec:related_work}


\header{Tabular error detection} Detecting errors in tabular data is a long-standing research problem in the data management community~\cite{chu2013discovering, abedjan2016detecting}. HoloDetect~\cite{rekatsinas2017holoclean} uses data augmentation and
few-shot learning to detect errors, while HoloClean~\cite{rekatsinas2017holoclean} uses probabilistic inference to find the best repair.
Raha~\cite{mahdavi2019raha} and Baran~\cite{mahdavi2020baran} leverage an ensemble of existing detectors, rules, and constraints for error detection and correction, respectively. 
ActiveClean~\cite{krishnan2016activeclean} applies active learning to iteratively repair the data while preserving monotone convergence guarantees.

Tabular error detection methods are designed to catch inconsistencies between columns but do not incorporate cross-modality information and may therefore struggle to detect multi-modal errors.

\header{Label error detection} Our problem can also be treated as label error detection, where one or more modalities are used to predict errors in a ``label column'' derived from the table.
For example, given the characteristics of e-Commerce products (from a table) and respective product images, we can use the images as input and one of the columns in the table as a label. 
A popular approach to label error detection is Cleanlab~\cite{northcutt2021confident, northcutt2021pervasive}.
Cleanlab leverages confident learning to improve existing models and estimate dataset problems such as erroneous labels, (near) duplicates, and non-IID data.
Second, \citeauthor{pmlr-v238-jager24a}~(\citeyear{pmlr-v238-jager24a}) proposed to apply conformal learning, a method to quantify and calibrate the uncertainty of ML models, for data cleaning.
Third, Data Shapley values~\cite{ghorbani2019data, karlavs2023data, wang2023data} are a data valuation metric to quantify the impact of each training point on a model's predictions, which has been shown to work well for label error detection. Fourth, the LEMoN~\cite{zhang2024lemonlabelerrordetection} framework leverages contrastive learning and a CLIP~\cite{clip} model. LEMoN finds the nearest neighbors of a sample on the image manifold and compares them to the neighbors on the textual manifold. 

However, label error detection methods are designed for predictive problems and concentrate on label errors, not covering, for instance, multi-column errors where several cells in a tuple contain correlated errors.
Moreover, it is costly to treat individual columns as labels as this often requires training a specialized model per column.

\header{Error detection with large language models} Recently, LLMs emerged as a powerful tool that can be prompted to detect errors in text, images, and structural data. FineMatch~\cite{Hang24_FineMatch} introduces a benchmark focusing on mismatch detection and correction in text-image pairs. FineMatch shows the proficiency of visual language models (VLMs), e.g., LLaVA~\cite{Haotian23_LLaVA} and GPT-4V~\cite{2023GPT4VisionSC}, in detecting and fixing errors in multi-modal inputs.
Versatile Data Cleanser~(VDC)~\cite{zhu2024vdc} is another LLM-powered label error detection framework that consists of three parts: Question generation, answering, and evaluation. Given an image and a textual label, VDC creates LLM-generated label-specific questions that are later answered by the multi-modal LLM based on the image. The visual question-answering and original labels are used to evaluate the correctness of labels.
Another LLM-based solution is DataVinci~\cite{Singh25_DataVinci}, which targets detecting and correcting sub-string errors.

While prior work has shown the effectiveness of LLMs and VLMs for text and image cleaning, it is still, to the best of our knowledge, unclear how VLMs handle tabular data and inter-row dependencies combined with visual data. Furthermore, VLMs maybe prohibitively expensive in settings with millions of input samples and hundreds of columns, especially since existing methods require several calls per sample to find erroneous data.

\section{Preliminary Experimental Results}
\label{sec:experiments}

Next, we conduct a set of preliminary experiments. We aim to show that multiple modalities indeed help to detect cross-modal errors and that existing baseline techniques do not sufficiently address our problem.

\subsection{Data and Error Generation}

\header{Datasets} We experiment with four datasets to analyze and demonstrate the difficulty of cross-modal error detection. All datasets contain tabular data and an image for each row in the table.
\textit{Fashion}~\cite{Data_Fashion} and \textit{Fashion 44K}~\cite{param_aggarwal_2019} are two similar Kaggle e-Commerce datasets with images and tabular data of clothing products, where \textit{Fashion 44K} is a larger version. The other two datasets, \textit{Baby} and \textit{Sports}, originate from subcategories of the Kaggle e-Commerce image dataset~\cite{Data_ecommerce}. They originally contain only images, and, therefore, we leverage a VLM (LLaVA 1.5-7b)~\cite{Haotian23_LLaVA} to generate corresponding tabular data (see prompts in \cref{sec:TableGen}), which we  manually post-process and refine to obtain a ground truth dataset.
All datasets contain an e-Commerce product title, category, type, and color attributes. There are also dataset-specific columns, e.g., sport type for the \textit{Sports} dataset. \Cref{tab:Datasets} and \cref{tab:Stats} in the Appendix highlight the properties of the datasets.

\begin{table*}[t!]
    \centering
    \caption{Error detection performance of the baseline approaches with varying modalities on three e-commerce different methods. Best result per dataset are highlighted in bold, the second-best underlined. We report some of the metrics for \method{LLaVA-Interleave (LLaVA-I.) (few-shot)} in brackets, where it only outputs a single class prediction (marking everything as erroneous). The best-performing methods are AutoGluon + DataScope and AutoGluon + Cleanlab with access to both image and tabular data. Runtimes are included in \cref{tab:FashionTimes}.}
    \vspace{0.3cm}
    \setlength\tabcolsep{1.2pt}
    \begin{tabular}{l|cc|ccc|ccc|ccc|ccc}
        \toprule
            & \textbf{Table} & \textbf{Image}& \multicolumn{3}{c|}{\textit{Fashion}} & \multicolumn{3}{c|}{\textit{Baby}} & \multicolumn{3}{c|}{\textit{Sports}} & \multicolumn{3}{c}{\textit{Fashion 44K}} \\
         \textbf{Method}  & used? & used? & $\mathcal{P}$ & $\mathcal{R}$ & $\mathcal{F}1$ & $\mathcal{P}$ & $\mathcal{R}$ & $\mathcal{F}1$ & $\mathcal{P}$ & $\mathcal{R}$ & $\mathcal{F}1$ & $\mathcal{P}$ & $\mathcal{R}$ & $\mathcal{F}1$ \\ \midrule
            \method{Raha}     & \checkmark & \xmark & \acc{0.16} & \acc{0.07} & \acc{0.09} & \acc{0.34} & \acc{0.15} & \acc{0.20} & \acc{0.17} & \acc{0.08} & \acc{0.10} & \acc{0.41} & \acc{0.29} & \acc{0.34}\\
            \method{AutoGluon + Cleanlab} & \checkmark & \xmark  & \acc{0.62} & \acc{0.41} & \acc{0.49} & \acc{0.48} & \acc{0.48} & \acc{0.48} & \acc{0.46} & \acc{0.57} & \acc{0.51} & \acc{0.68} & \acc{0.29} & \acc{0.41} \\
            \method{AutoGluon + DataScope} & \checkmark & \xmark & \acc{0.37} & \acc{0.78} & \acc{0.50} & \acc{0.55} & \acc{0.74} & \acc{0.63} & \acc{0.42} & \acc{0.66} & \acc{0.52} & \acc{0.37} & \acc{0.81} & \acc{0.51} \\
            \method{LLaVA (zero-shot)} & \checkmark & \xmark & \acc{1.00} & \acc{0.001} & \acc{0.003} & \acc{0.71} & \acc{0.14} & \acc{0.23} & \acc{0.84} & \acc{0.37} & \acc{0.52} & \acc{0.94} & \acc{0.00} & \acc{0.01}\\ 
            \method{LLaVA (few-shot)} & \checkmark & \xmark & \acc{0.71} & \acc{0.34} & \acc{0.46} & \acc{(0.50)} & \acc{(1.00)} & \acc{(0.67)} & \acc{(0.50)} & \acc{(1.00)} & \acc{(0.67)} & \acc{0.50} & \acc{0.71} & \acc{0.59} \\ 
            
            \midrule
            \method{AutoGluon + Cleanlab} & \xmark & \checkmark & \acc{0.71} & \acc{0.94} & \acc{0.81} & \acc{0.65} & \acc{0.66} & \acc{0.66} & \acc{0.61} & \acc{0.70} & \acc{0.65} & \acc{0.56} & \acc{0.92} & \underline{\acc{0.70}}\\
            \method{AutoGluon + DataScope} & \xmark &\checkmark & \acc{0.82} & \acc{0.96} & \textbf{\acc{0.89}} & \acc{0.81} & \acc{0.96} & \underline{\acc{0.88}} & \acc{0.67} & \acc{0.74} & \underline{\acc{0.70}} & \acc{0.53} & \acc{0.96} & \acc{0.68} \\ 

            \method{LLaVA (zero-shot)} & \xmark & \checkmark   & \acc{0.03} & \acc{0.03} & \acc{0.03} & \acc{0.07} & \acc{0.06} & \acc{0.06} & \acc{050} & \acc{0.005} & \acc{0.01} & \acc{0.008} & \acc{0.00} & \acc{0.00}\\  
            \method{LLaVA-I. (few-shot)} & \xmark & \checkmark & \acc{0.15} & \acc{0.98} & \acc{0.27} & \acc{0.17} & \acc{0.64} & \acc{0.27} & \acc{0.26} & \acc{0.94} & \acc{0.41} & \acc{0.11} & \acc{0.99} & \acc{0.21}\\  

          \midrule
            \method{AutoGluon + Cleanlab} & \checkmark & \checkmark & \acc{0.87} & \acc{0.78} & \underline{\acc{0.83}} & \acc{0.73} & \acc{0.67} & \acc{0.70} & \acc{0.62} & \acc{0.70} & \acc{0.66} & \acc{0.80} & \acc{0.71} & \textbf{\acc{0.75}}\\
            \method{AutoGluon + DataScope} & \checkmark & \checkmark & \acc{0.83} & \acc{0.82} & \underline{\acc{0.83}} & \acc{0.84} & \acc{0.93} & \textbf{\acc{0.89}} & \acc{0.72} & \acc{0.75} & \textbf{\acc{0.73}} & \acc{0.75} & \acc{0.47} & \acc{0.58} \\
           \method{LLaVA (zero-shot)} & \checkmark & \checkmark & \acc{0.00} & \acc{0.00} & \acc{0.00} & \acc{1.00} & \acc{0.006} & \acc{0.01} & \acc{0.00} & \acc{0.00} & \acc{0.00} & \acc{0.00} & \acc{0.00} & \acc{0.00}\\  
           \method{LLaVA-I. (few-shot)} & \checkmark & \checkmark & \acc{(0.50)} & \acc{(1.00)} & \acc{(0.67)} & \acc{0.51} & \acc{0.95} & \acc{0.62} & \acc{(0.50)} & \acc{(1.00)} & \acc{(0.67)} & \acc{(0.49)} & \acc{(0.99)} & \acc{(0.66)}\\ 
           \method{LEMoN} & \checkmark & \checkmark & \acc{0.72} & \acc{0.41} & \acc{0.52} & \acc{0.66} & \acc{0.37} & \acc{0.48} & \acc{0.51} & \acc{0.36} & \acc{0.42} & \acc{0.50} & \acc{0.40} & \acc{0.44}\\

         \bottomrule
    \end{tabular}
    \vspace{-0.2cm}
    
    \label{tab:baseline-performance}
    \vspace{-0.3cm}
    
\end{table*}

\header{Error injection} Inspired by our interview with practitioners, we inject synthetic errors into the test splits of the tabular data for our datasets to simulate cross-modal errors that are hard to detect from one modality alone. Note that none of the training splits contain errors. First, we manually curate the data to remove pre-existing errors by inspecting the data and running Cleanlab with the original non-corrupted tabular data to find and fix inconsistencies. Next, we randomly select 50\% of the rows of each test split. For each sampled row of the test data, we select a random column to introduce the error into. To inject an error, we replace the selected cell with a random existing value from the set of column unique values different from the cell's current value. In addition, we modify all cells in the selected row which contain the original cell value. For instance, if we change the color attribute of a product, we also replace the name of the color in the product title if contained. By that, we ensure that the error must be detected from the image and that the original value is not leaked from other cell values.
For correlated columns, we replace the original values only with already observed pairs, e.g., to avoid creating non-sensical products with the category ``footwear'' and subcategory ``dress''.

\subsection{Baseline Error Detection Performance}

The goal of our first experiment is two-fold: we investigate whether the image modality helps with finding cross-modal errors in tabular data, and we assess the performance of several baselines on our benchmark data.

\header{Experimental setup} We evaluate the error detection performance of several baseline approaches from \cref{sec:related_work} with different modalities (table only, image only, table + image). For \textit{Fashion} and \textit{Fashion 44K}, we use a random sample of 30\% of the tuples as test set, for \textit{Baby} and \textit{Sports}, we use the union of their existing validation and test splits as test set. As discussed, we only introduce synthetic errors into the test data. We measure the performance with precision ($\mathcal{P}$), recall ($\mathcal{R}$), and F1 score ($\mathcal{F}1$).

\header{Methods} We evaluate the following methods in our benchmark and refer to~\cref{sec:prompts} for details on used prompts.

\begin{itemize}[itemsep=0.2em,leftmargin=*]
  \item \textbf{\method{Raha}} -- a state-of-the-art single-column error detection framework for tabular data~\cite{mahdavi2019raha}. We use the original benchmarking scripts and provide the framework with clean and dirty data samples. Raha does not support images, therefore, we evaluate Raha only with tabular data.

  \item \textbf{\method{AutoGluon + Cleanlab}} -- we combine the state-of-the-art AutoML library AutoGluon~\cite{tang2024autogluon}, with a state-of-the-art method for label error detection and correction Cleanlab~\cite{northcutt2021confident, northcutt2021pervasive}. We repeatedly train AutoGluon models with each column as a target and use the resulting classifiers as input for Cleanlab's error detection. We vary the input modalities for the AutoGluon models from table-only to image-only and table combined with image. 

  \item \textbf{\method{AutoGluon + DataScope}} -- we combine AutoGluon with the DataScope library~\cite{karlavs2023data} to identify erroneous tuples via negative data importance scores. Concretely, we compute Data Shapley values~\cite{ghorbani2019data} for the potentially dirty test data, using the clean training data for validation. We configure DataScope to calculate exact Data Shapley values for a kNN proxy model~\cite{jia12efficient} with $k=1$, repeatedly train AutoGluon models with each column as a target and use the resulting feature representations as input for DataScope. 
  We apply DataScope to the erroneous test set and score on clean train data. All instances with negative importance are considered erroreneous.
 
  \item \textbf{\method{LLaVA}} -- we prompt the vision language models LLaVA-1.5 7b~\cite{Haotian23_LLaVA} and LLaVA-Next-Interleave 7b~\cite{li2024llavanextinterleavetacklingmultiimagevideo} to detect cross-modal errors.
  For the zero-shot variant, we do not provide any examples, while we add up to ten concrete dataset- and modality-specific examples (image and label/table row) to the prompt for the few-shot variant.
  We use LLaVA-Next-Interleave for the few-shot setting since LLaVA-1.5 does not support multi-image inference.
  
  \item \textbf{\method{LEMoN}} -- to evaluate the potential of contrastive learning for cross-modal error detection, we leverage LEMoN~\cite{zhang2024lemonlabelerrordetection} that can handle textual labels only. We create labels from tabular data by serializing each row into a string. The final label is a combination of column names and values, e.g., for \textit{Fashion} data an example label text is \textit{ProductTitle - Nike Men Air Zoom Shoes, Gender - Men, Category - Footwear...}. LEMoN requires hyperparameters like noise type and noise level; we use random noise with a level of \num{0.4}, analogous to the settings in the original paper. 
\end{itemize}

\subsection{Results and Discussion} 

\header{Baseline performance} We show the performance scores for the baselines in \Cref{tab:baseline-performance}, together with an indication of the modalities used for the predictions. First, as expected, we observe that the cross-modal errors are hard to detect from the tabular data alone, indicated by the low scores of \method{Raha}, a state-of-the-art error detector for tabular data. Second, we see sub-par results for the other methods as well, when they only have access to the tabular data. Third, the image modality helps with error detection, and the $\mathcal{F}1$ scores of \method{AutoGluon + Cleanlab} and \method{AutoGluon + DataScope} in the image-only setup are significantly higher than in the table-only setup. Furthermore, in three out of four datasets, joint access to both modalities results in the best performance, with an improvement of up to 5\% in $\mathcal{F}1$ score compared to the image-only setup. However, for \textit{Fashion}, the performance of \method{AutoGluon + DataScope} degrades when images are combined with tabular data.

The results confirm our hypothesis that the chosen problem is difficult, even for VLMs. \method{LLaVA} in both modes, zero- and few-shot, produces low or unreliable $\mathcal{F}1$ scores and, in several cases, even marks every input tuple as erroneous (shown in brackets). This is surprising since we use LLaVA for the tabular data generation of the \textit{Baby} and \textit{Sports} datasets.

\begin{table}[!t]
    \centering
    \caption{\centering Examples of cross-modal errors that are detectable by looking at table, image, and jointly image and tabular data.}
    \vspace{0.3cm}
    \setlength\tabcolsep{3.5pt}
    \begin{tabular}{c|c|c}
    \toprule
     \multirow{2}{*}{\textbf{Erroneous Tuple}} & \multirow{2}{*}{\textbf{Image}} & \textbf{How To} \\
      &  & \textbf{Detect} \\
    \midrule
        \begin{tabular}{r|l}
         Category & \emph{\color{red}{Bags}}\\
         ProductType & \emph{\color{red}{Trolley Bag}}\\
         Color & White\\
         ProductTitle & Puma ... Shoe\\
         \end{tabular}
         &  \raisebox{-2.7em}{\transparent{0.4}\includegraphics[height=2cm]{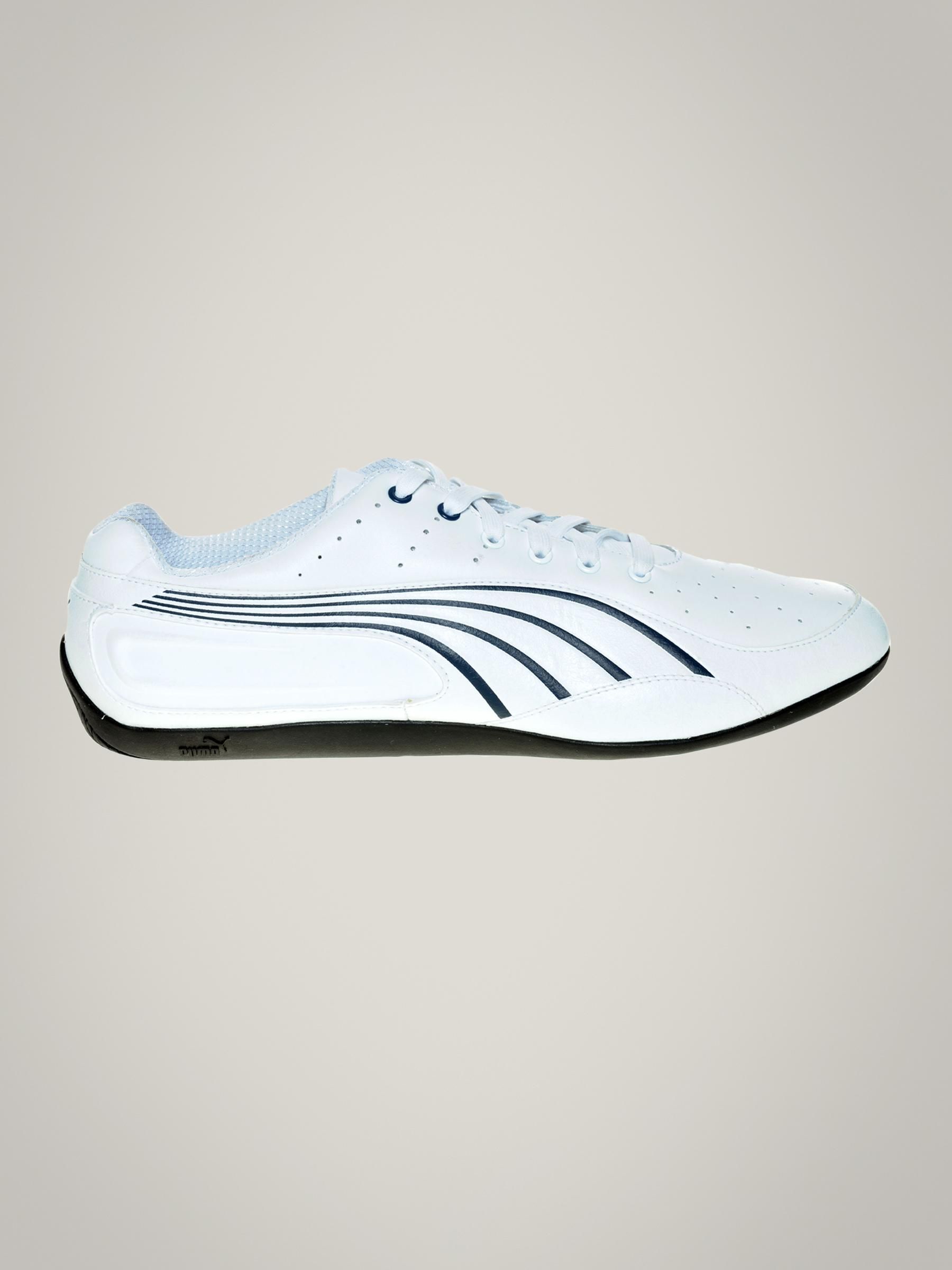}} 
         & \begin{tabular}{c}
         Table \\
         \end{tabular}\\ 
         \midrule
         \midrule
    
        \begin{tabular}{r|p{2cm}}
         \text{\color{gray}Category} & \text{\color{gray}Apparel} \\
         \color{gray}{ProductType} & \text{\color{gray}Top}\\
         Gender & \emph{\color{red}{Boys}}\\
         \color{gray}{Color} & \text{\color{gray}Purple}\\
         \end{tabular}
         &  \raisebox{-2.7em}{\includegraphics[height=2cm]{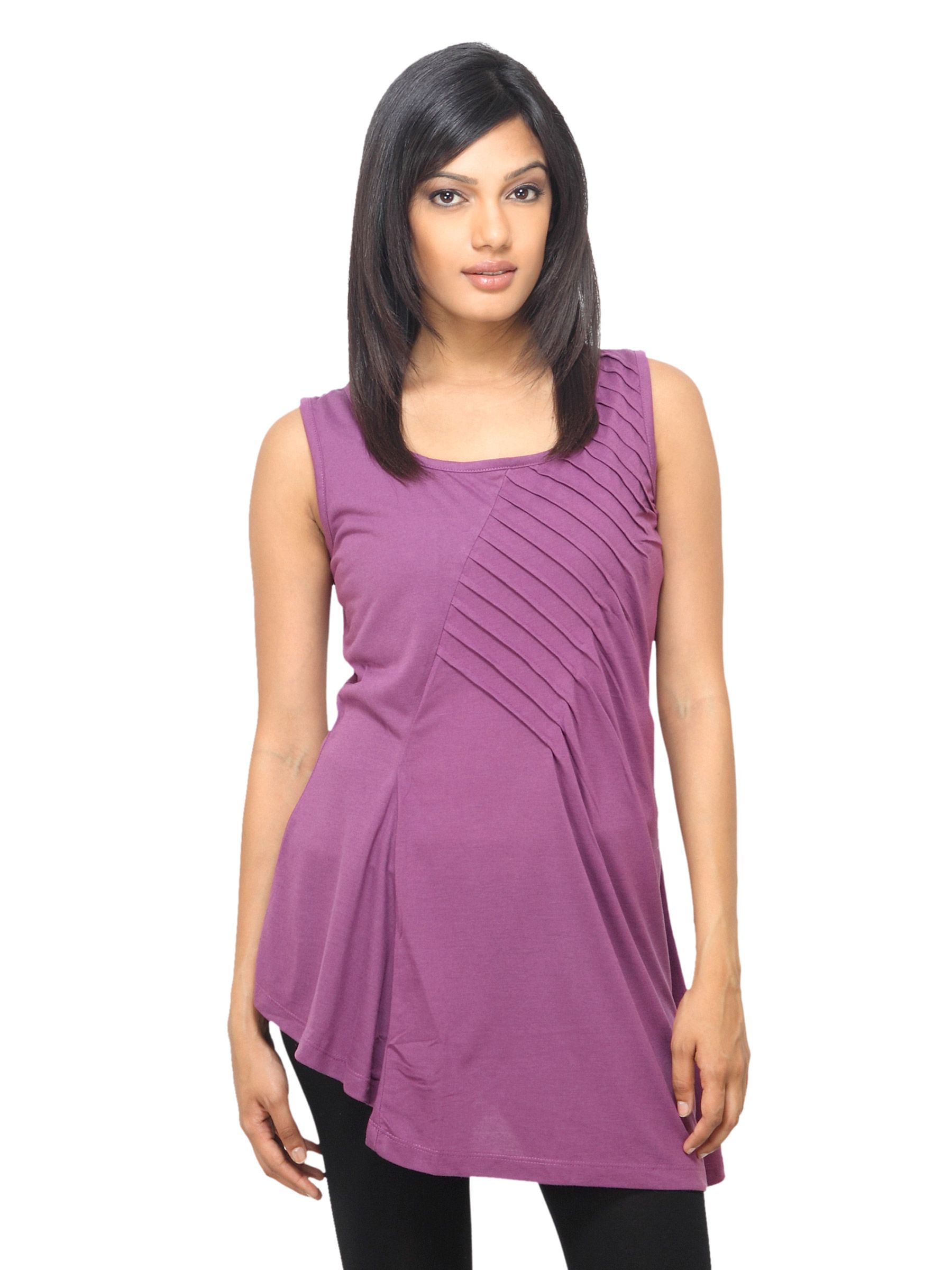}} 
         & \begin{tabular}{c}
         Image \\
         \end{tabular}\\ 
         \midrule
         \midrule
         
         \begin{tabular}{r|p{2cm}}
         Category & \emph{\color{red}{Clothing}}\\
         ProductType & \emph{\color{red}{Pants}}\\
         Color & Blue pink\\
         Material & Plastic \\
         \end{tabular}
         &  \raisebox{-2.7em}{\includegraphics[height=2cm]{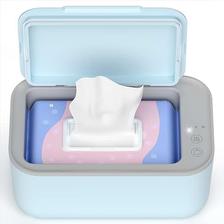}} 
         & \begin{tabular}{c}
         Table \\
         + \\
         Image \\
         \end{tabular}\\ 
         \midrule
         \begin{tabular}{r|l}
         Category & \emph{\color{red}{Volleyball}}\\
         ProductType & \emph{\color{red}{Volleyball nets}}\\
         Color & Black\\
         SportType & Camping \\
         \end{tabular} & 
         \raisebox{-2.7em}{\includegraphics[height=2cm]{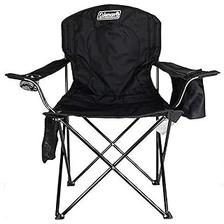}} 
         & \begin{tabular}{c}
         Table \\
         + \\
         Image \\
         \end{tabular}\\
         \bottomrule
    \end{tabular}
    \vspace{-0.2cm}
    
    \label{tab:ExampleProducts}

    \vspace{-0.3cm}
\end{table}

\begin{table*}[ht]
    \centering
    \caption{Selection of easy and hard columns for cross-modal error detection using AutoGluon + Cleanlab and AutoGluon + DataScope. Hard columns have higher number of distinct values and a more skewed frequency distribution. 
    }
    \vspace{0.3cm}
    \setlength\tabcolsep{2.6pt}
    \begin{tabular}{l|ll|l|ccc|c|c}
        \toprule
         & \multirow{2}{*}{\textbf{Dataset}} & \multirow{2}{*}{\textbf{Column}} &  \multirow{2}{*}{\textbf{Method}} &  $\mathcal{F}1$ & $\mathcal{F}1$ & $\mathcal{F}1$ & \textbf{Cardi-} & \multirow{2}{*}{\textbf{Frequency Distribution}} \\
         & & & & Table & Image & Table + Image & \textbf{nality} & \\ \midrule
         
        \multirow{6}{*}{\rotatebox{90}{\textbf{Easy}}} 
            & \multirow{2}{*}{\textit{Fashion}} & \multirow{2}{*}{Category} & Cleanlab & 0.48 & 0.94 & \textbf{1.00} & \multirow{2}{*}{2} & \multirow{2}{*}{\includegraphics[trim=0 0 75 0,clip]{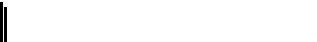}} \\\vspace{0.1cm} 
            ~    & & & DataScope & 0.43 & \textbf{1.00} & \textbf{1.00} & \\ 
        ~   & \multirow{2}{*}{\textit{Fashion}} & \multirow{2}{*}{SubCategory} & Cleanlab & 0.61 & \textbf{0.94} & \textbf{0.94} & \multirow{2}{*}{9} & \multirow{2}{*}{\includegraphics[trim=0 0 75 0,clip]{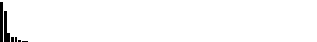}} \\\vspace{0.1cm}  
        ~    & & & DataScope & 0.47 & 0.87 & 0.91 & \\
        ~   & \multirow{2}{*}{\textit{Baby}} & \multirow{2}{*}{PackageMaterial} & Cleanlab& 0.70 & 0.89 & 0.90 & \multirow{2}{*}{9} & \multirow{2}{*}{\includegraphics[trim=0 0 75 0,clip]{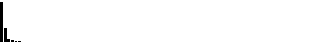}} \\\vspace{0.1cm}  
        ~    & & & DataScope & 0.70 & 0.94 & \textbf{0.95} & \\

        \midrule
        \multirow{6}{*}{\rotatebox{90}{\textbf{Hard}}} 
        ~    & \multirow{2}{*}{\textit{Fashion}} & \multirow{2}{*}{Color} & Cleanlab ~    & 0.58 & 0.66 & 0.69 & \multirow{2}{*}{38} & \multirow{2}{*}{\includegraphics[trim=0 0 75 0,clip]{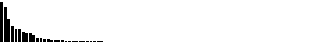}} \\\vspace{0.1cm}  
        ~    & & & DataScope & 0.51 & \textbf{0.77} & 0.71 & \\
        ~    & \multirow{2}{*}{\textit{Sports}} & \multirow{2}{*}{SportType} & Cleanlab & 0.59 & 0.59 & 0.62 & \multirow{2}{*}{65} & \multirow{2}{*}{\includegraphics[trim=0 0 75 0,clip]{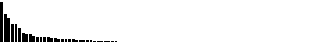}} \\\vspace{0.1cm}
        ~    & & & DataScope & 0.47 & \textbf{0.70} & 0.68 & \\
        ~    & \multirow{2}{*}{\textit{Baby}} & \multirow{2}{*}{ProductType} & Cleanlab & 0.43 & 0.42 & 0.51 & \multirow{2}{*}{132} & \multirow{2}{*}{\includegraphics[trim=0 0 75 0,clip]
        {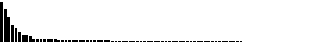}} \\
        ~    & & & DataScope & 0.70 & \textbf{0.88} & \textbf{0.88} & \\  

        \bottomrule
    \end{tabular}
    \vspace{-0.2cm}
    
    \label{tab:ColumnPerformance}
\end{table*}

Even though \method{AutoGluon + DataScope} scores the highest in our benchmark, it only reaches high $\mathcal{F}1$ for the two small datasets and provides subpar performance on the larger \textit{Fashion44K} dataset, missing up to half of the errors in some cases (as indicated by the recall scores).
The inconsistent performance across modalities and datasets raises the question of when tabular data combined with images becomes helpful for DataScope, given that the multi-modal approach has proven effective for other methods.
Furthermore, due to the use of a kNN proxy model, DataScope's performance relies heavily on AutoGluon's input embedding quality, with $\mathcal{F}1$ scores decreasing by 10\% when using insufficiently trained representations (not shown in the table).
Furthermore, DataScope requires a large clean validation set. While our benchmark uses clean training data for scoring, this requirement is problematic in real-world applications where such clean datasets may not be available.
\method{AutoGluon + Cleanlab} scores the second highest in our benchmark but only reaches $\mathcal{F}1$ scores of around 70\%-80\% percent, missing 22\% to 33\% percent of errors (according to the recall scores). Interestingly, we see a consistent positive impact of having joint access to both modalities (table + image) for this baseline, with improvements of up to 5\% in F1 scores.

We interpret our results as confirmation that further research is needed for cross-modal error detection where one of the modalities is a table.


\headerl{Example} To give a concrete example of cross-modal errors that are found only by jointly looking at the image and table, we point to \cref{tab:ExampleProducts}, which describes two cross-modal examples from the \textit{Baby} and \textit{Sports} datasets together with errors that are detectable from a single modality from the \textit{Fashion} dataset. The baby wipes have the wrong product type and category, and are described as baby pants in the table. The camping chair is wrongly marked as a volleyball net. In both cases, the table-only or image-only methods fail to detect these inconsistencies. 
Both errors are nontrivial. While the ``wipes'' error happens even during data generation, where LLaVA confuses baby wipes/onesies/diapers, the ``chair'' error is detected only when given both image and table which we attribute to the fact that the sport type contradicts the category and helps to detect the inconsistency.

\header{Column-wise performance} To deeper understand the results of the best-performing methods \method{AutoGluon + DataScope} and \method{AutoGluon + Cleanlab}, we analyze their performance for errors in different columns. \Cref{tab:ColumnPerformance} shows three easy (high $\mathcal{F}1$ score) and the three hard (low $\mathcal{F}1$ score) cases. 
Overall, the analysis indicates that it is more difficult to detect errors in high-cardinality columns (many distinct values) with a skewed frequency distribution (many rare values), e.g., the product type column in \textit{Baby} is hard to detect because of a long-tailed distribution with only a few common values.
Importantly, real-world data often exhibit skewed/long-tailed distributions~\cite{yi2025geometry}. On the other hand, \cref{tab:ColumnPerformance} indicates that errors are the easiest to detect in columns with few distinct values and balanced frequencies.
An key observation, confirmed by the column-wise evaluation, is that joint access to image and table data improves the $\mathcal{F}1$ scores.
However, DataScope struggles to leverage tabular data with images in challenging cases. 
The optimal approach for combining modalities remains unclear for DataScope, while other methods successfully benefit from multi-modal error detection.

\begin{table}[t!]
    \setlength\tabcolsep{5.4pt}
    \centering
    \caption{A selection of easy and hard columns for error detection and repair using AutoGluon + Cleanlab. The table contains the error correction \emph{accuracy} per column. 
    }
    \vspace{0.3cm}
    \begin{tabular}{c|l|l|c|c|c}
        \toprule
        & \multirow{2}{*}{\textbf{Dataset}} & \multirow{2}{*}{\textbf{Column}} & \multirow{2}{*}{Table} & \multirow{2}{*}{Image} & Table+ \\
        & & & & & Image\\
        \midrule
        \multirow{3}{*}{\rotatebox{90}{\textbf{Easy}}}&\textit{Fashion} & Sub      & \acc{0.01} & \acc{0.25} & \textbf{\acc{0.83}} \\
        &\textit{Baby} & PackageMat.   & \acc{0.79} & \acc{0.18} & \textbf{\acc{0.81}} \\
        &\textit{Fashion} & SubCategory & \acc{0.07} & \acc{0.66} & \textbf{\acc{0.94}} \\

        \midrule

        \multirow{3}{*}{\rotatebox{90}{\textbf{Hard}}}&\textit{Baby} & ProductType     & \acc{0.17} & \acc{0.20} & \textbf{\acc{0.24}} \\
        &\textit{Baby} & Color           & \textbf{\acc{0.21}} & \acc{0.16} & \textbf{\acc{0.21}} \\
        &\textit{Sports} & ProductType   & \acc{0.05} & \acc{0.30} & \textbf{\acc{0.35}} \\
        
        \bottomrule
    \end{tabular}
    \vspace{-0.5cm}
    
    \label{tab:SpecifcCategoryCorrectionAll}
\end{table}

\subsection{Automated Repair} 
While this paper focuses on error detection, an important next step is the automated repair of the detected errors. 
For that, we evaluate AutoGluon + Cleanlab's ability to correct the errors by leveraging the correct ``label'' as suggested via confident learning. 
\cref{tab:SpecifcCategoryCorrectionAll} shows the error detection and correction accuracy for the selected columns in all three datasets, denoting the fraction of injected errors that could be successfully detected and repaired. Full results are in Appendix, \cref{tab:SpecifcCategoryCorrectionFashion}, \cref{tab:SpecifcCategoryCorrectionBaby}, \cref{tab:SpecifcCategoryCorrectionSports}, and \cref{tab:SpecifcCategoryCorrectionFashionBig}. 
Similar to detection, we observe that error correction benefits from joint access to tabular and image data and that repairs are more difficult for heavy-tailed data (e.g., in color and product type from the \textit{Baby} and \textit{Sports} datasets). There is room for improvement in error detection and repair. 

\section{Conclusions}
\label{sec:next}

Our preliminary findings highlight a gap in existing methodologies: the absence of an effective approach that can directly handle multi-modal inputs for both error detection and correction. We found tabular error detection methods and VLMs to have subpar performance. While label error detection methods showed potential, they still incur a performance gap, and it is unclear how to select the best approach among them. Furthermore, label error detection methods are expensive, as one has to train a separate AutoML model for each column in every dataset. 

To advance cross-modal error detection for tables, we plan to design a dedicated multi-modal model (potentially based on self-supervised contrastive learning~\cite{clip}) and to develop a larger and more comprehensive benchmark that includes real-world data and cross-modal errors from domains beyond e-Commerce. Additionally, a broader range of baseline models should be considered in the evaluation, e.g., tabular foundation models like TabPFN~\cite{Hollmann2025} and CARTE~\cite{Kim2024_CARTE}, as well as novelty and anomaly detection methods. 

\section*{Acknowledgements}

The authors thank Sebastian Baunsgaard and 
Bojan Karlaš for their insightful comments and constructive feedback on the manuscript.

\section*{Impact Statement}


This work introduces the problem of cross-modal error detection in tabular data and benchmarks current approaches to advance data preparation and validation. 
As enterprises increasingly leverage multi-modal datasets—combining structured tables with accompanying text and images—ensuring consistency across modalities becomes vital. However, existing methods focus mainly on single-modality or label errors, often missing cross-modal inconsistencies common in domains like e-Commerce and healthcare.
Such inconsistencies can lead to serious real-world consequences, including safety risks for consumers, for example, a product image showing a adults-only product mislabeled as child-safe, or a violent movie tagged as family-friendly. Moreover, undetected errors can result in violations of legal and regulatory standards, such as incorrect specification of allergens, age restrictions, or chemical contents.
Our aim is to lay the groundwork for modality-aware data validation pipelines that not only enhance technical robustness but also promote consumer protection and regulatory compliance.

\balance
\bibliography{refs}
\bibliographystyle{icml2025}

\onecolumn
\newpage
\appendix

\section{Appendix}


\subsection{Tabular Data Generation}
\label{sec:TableGen}

The prompt used to generate the tabular data for \textit{Baby} is:

\begin{lstlisting}[style=promtstyle]
Given this image of baby product, fill these attributes of a table: {",".join(category_fields)}. Use only basic colors. The Age Limit should be a number. Title should be a combination of {",".join(category_fields_in_title))}. Return the result as a JSON with the attributes.
\end{lstlisting}

The prompt used to generate the tabular data for \textit{Sports} is:

\begin{lstlisting}[style=promtstyle]
Given this image of products for sport/outdoor activities, fill these attributes of a table: {",".join(category_fields)}. Use only basic colors. Title should be a combination of {",".join(category_fields_in_title))}. Return the result as a JSON with the attributes.
\end{lstlisting}

\subsection{LLM Prompting for Error Detection}
\label{sec:prompts}

To evaluate LLaVA 1.5-7b~\cite{Haotian23_LLaVA} with a single table, we prompt LLM with each row from the table:

\begin{lstlisting}[style=promtstyle]
Given a set of e-Commerce product properties, answer if there are errors in the product properties.
Properties: {col_name-row_value-pairs}.
Please only answer with 'yes' if there are errors, or 'no' if there are no errors.
\end{lstlisting}

The prompts used to assess performance of the LLM with images and a single value from the label column:

\begin{lstlisting}[style=promtstyle]
Given am e-Commerce product image and property, answer if the property is erroneous, especially comparing to the image.
Property: {col_name-row_name}.
Please only answer with 'yes' if there are errors, or 'no' if there are no errors.
\end{lstlisting}

The prompts used to assess performance of the LLM with images and tabular data:

\begin{lstlisting}[style=promtstyle]
Given an e-Commerce product image and set of product properties, answer if there are inconsistencies between product properties and the image. 
Properties: {col_name-row_name-pairs}.
Please only answer with 'yes' if there are errors, or 'no' if there are no errors.
\end{lstlisting}

For LLaVA-Next Interleave 7b~\cite{li2024llavanextinterleavetacklingmultiimagevideo}, we add additional examples with images and expected result to the prompts above.
For example, for images with a single attribute as a target, we use:

\begin{lstlisting}[style=promtstyle]
<|im_start|>user \nGiven an e-Commerce product image and property, answer if the product property contains errors, especially comparing to the image. Here are some examples: 
    
Example 1: Image: <image>. Properties: Category - care. Assessment: yes. Category is wrong. Category should be diapers. 
Example 2: Image: <image>. Properties: Product type - wipes. Assessment: yes. Product type is wrong. Product type should be diapers. 
Example 3: Image: <image>. Properties: Color - multi-colored. Assessment: no.
...

Please evaluate a product with image <image> and the following property: {col_name-row_name}. Please only answer with 'yes' if there are errors, or 'no' if there are no errors.|im_end|><|im_start|>assistant 
\end{lstlisting}

\newpage

\subsection{Supplementary Tables}

\setcounter{table}{0}
\renewcommand{\thetable}{A\arabic{table}}

\begin{table}[H]
    \centering
    \caption{Basic statistics of the multi-modal e-Commerce datasets.}
    \vspace{0.1cm}
    
    \setlength\tabcolsep{7.7pt}
    \begin{tabular}{l|cccc}
        \toprule
        \textbf{Name}   & \textbf{\#Rows} & \textbf{\#Cols} & \textbf{Image size}\\ \midrule
        \textit{Fashion}& \numprint{2907} & 6 & 1080\raisebox{0.07em}{$\times$}1440 \\ 
        \textit{Baby}   & \numprint{1299} & 8 & 224\raisebox{0.07em}{$\times$}224 \\
        \textit{Sports} & \numprint{1368} & 7 & 224\raisebox{0.07em}{$\times$}224\\
        \textit{Fashion 44K} & \numprint{44442} & 9 & 1080\raisebox{0.07em}{$\times$}1440\\
       \bottomrule 
    \end{tabular}
    
    \vspace{-0.2cm}

    \vspace{-0.1cm}
    \label{tab:Datasets}
\end{table}

\begin{table*}[h]
    \centering
    \caption{Statistics of all columns and datasets, including value counts histogram of each column, sorted descending by counts.}
    \vspace{0.1cm}
    
    \setlength\tabcolsep{2.0pt}
    \begin{tabular}{c|r|c|c}
        \toprule
         \textbf{Dataset} & \textbf{Column} & \textbf{\#Distinct} & \textbf{Distinct Frequency Distribution} \\ \midrule
            \multirow{5}{*}{\raisebox{-3.5em}{\textit{Fashion}}} & \ri{Gender} & \ri{4} & \includegraphics[]{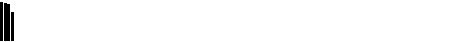} \\
            & \ri{Category} & \ri{2} & \includegraphics[]{figures/stats/Category_dirty_fashion.pdf} \\
            & \ri{SubCategory} & \ri{9} & \includegraphics[]{figures/stats/SubCategory_dirty_fashion.pdf} \\
            & \ri{ProductType} & \ri{30} & \includegraphics[]{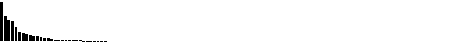} \\
            & \ri{Color} & \ri{38} & \includegraphics[]{figures/stats/Colour_dirty_fashion.pdf} \\

            \midrule
            \multirow{4}{*}{\raisebox{-2.5em}{\textit{Baby}}} & \ri{Category} & \ri{20} & \includegraphics[]{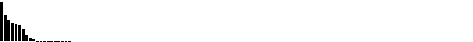} \\
            & \ri{ProductType} & \ri{132} & \includegraphics[]{figures/stats/GeneratedProductType_dirty_baby.pdf} \\
            & \ri{Color} & \ri{107} & \includegraphics[]{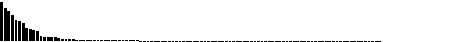} \\
            & \ri{PackageMaterial} & \ri{9} & \includegraphics[]{figures/stats/GeneratedPackageMaterial_dirty_baby.pdf} \\

            \midrule
            \multirow{4}{*}{\raisebox{-2.5em}{\textit{Sports}}} & \ri{Category} & \ri{55} & \includegraphics[]{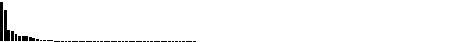} \\
            & \ri{ProductType} & \ri{344} & \includegraphics[]{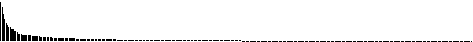} \\
            & \ri{Color} & \ri{77} & \includegraphics[]{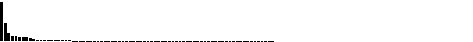} \\
            & \ri{SportType} & \ri{65} & \includegraphics[]{figures/stats/GeneratedSportType_dirty_sports.pdf} \\

            \midrule
            \multirow{6}{*}{\raisebox{-3.5em}{\textit{Fashion 44K}}} & \ri{Gender} & \ri{5} & \includegraphics[]{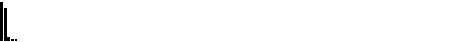} \\
            & \ri{Category} & \ri{7} & \includegraphics[]{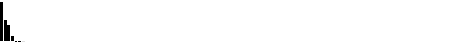} \\
            & \ri{SubCategory} & \ri{45} & \includegraphics[]{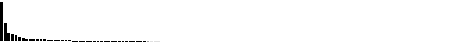} \\
            & \ri{ProductType} & \ri{144} & \includegraphics[]{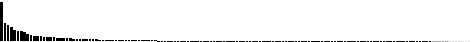} \\
            & \ri{Color} & \ri{47} & \includegraphics[]{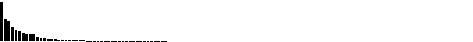} \\
            & \ri{Season} & \ri{5} & \includegraphics[]{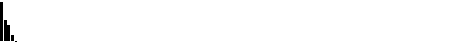} \\
            
         \bottomrule
    \end{tabular}
    \vspace{-0.2cm}
    
    \label{tab:Stats}
\end{table*}

\newpage

\begin{table}[H]

    \vspace{-0.5cm}
    
    \centering
    \caption{AutoGluon + Cleanlab performance per column in \textit{Fashion}.}
    \vspace{0.1cm}
    
    \setlength\tabcolsep{4.2pt}
    \begin{tabular}{c|c|c|c|c|c|c|c}
        \toprule
         \textbf{Table} & \textbf{Image} & \textbf{Measure} &  \textbf{Gender} & \textbf{Category} & \textbf{SubCategory} & \textbf{ProductType} & \textbf{Color} \\ \midrule

        \multirow{3}{*}{\checkmark} & \multirow{3}{*}{\xmark} & $\mathcal{P}$  
                               & \acc{1.00} & \acc{1.00} & \acc{0.69} & \acc{0.54} & \acc{0.61} \\
            & & $\mathcal{R}$  & \acc{0.01} & \acc{0.32} & \acc{0.55} & \acc{0.60} & \acc{0.55}\\
            & & $\mathcal{F}1$ & \acc{0.02} & \acc{0.48} & \acc{0.61} & \acc{0.57} & \acc{0.58}\\

        \midrule
        
        \multirow{3}{*}{\xmark} & \multirow{3}{*}{\checkmark} & $\mathcal{P}$  
                               & \acc{0.79} & \acc{1.00} & \acc{0.91} & \acc{0.66} & \acc{0.52} \\
            & & $\mathcal{R}$  & \acc{0.95} & \acc{0.88} & \acc{0.98} & \acc{0.93} & \acc{0.93}\\
            & & $\mathcal{F}1$ & \acc{\textbf{0.86}} & \acc{0.94} & \acc{\textbf{0.94}} & \acc{0.77} & \acc{0.66}\\

        \midrule
         
         \multirow{3}{*}{\checkmark} & \multirow{3}{*}{\checkmark} & $\mathcal{P}$  & \acc{0.95} & \acc{1.00} & \acc{0.94} & \acc{0.78} & \acc{0.75} \\
            & & $\mathcal{R}$  & \acc{0.57} & \acc{1.00} & \acc{0.94} & \acc{0.88} & \acc{0.64}\\
            & & $\mathcal{F}1$ & \acc{0.71} & \acc{\textbf{1.00}} & \acc{\textbf{0.94}} & \acc{\textbf{0.83}} & \acc{\textbf{0.69}}\\
         \bottomrule
    \end{tabular}
    \label{tab:SpecifcCategoryPerformanceFashion}
\end{table}

\begin{table}[H]

    \vspace{-0.7cm}
    
    \centering
    \caption{AutoGluon + Cleanlab performance per column in \textit{Baby}.}
    \vspace{0.1cm}
    
    \setlength\tabcolsep{5.8pt}
    \begin{tabular}{c|c|c|c|c|c|c}
        \toprule
         \textbf{Table} & \textbf{Image} & \textbf{Measure} & \textbf{Category} & \textbf{ProductType} & \textbf{Color} & \textbf{PackageMaterial} \\ \midrule

            \multirow{3}{*}{\checkmark} & \multirow{3}{*}{\xmark} 
            & $\mathcal{P}$    & \acc{0.70} & \acc{0.54} & \acc{0.26} & \acc{0.61} \\
            & & $\mathcal{R}$  & \acc{0.34} & \acc{0.36} & \acc{0.39} & \acc{0.81} \\
            & & $\mathcal{F}1$ & \acc{0.46} & \acc{0.43} & \acc{0.32} & \acc{0.70}\\

            \midrule
            \multirow{3}{*}{\xmark} & \multirow{3}{*}{\checkmark} 
            & $\mathcal{P}$    & \acc{0.70} & \acc{0.44} & \acc{0.55} & \acc{0.87} \\
            & & $\mathcal{R}$  & \acc{0.88} & \acc{0.41} & \acc{0.45} & \acc{0.92} \\
            & & $\mathcal{F}1$ & \acc{0.78} & \acc{0.42} & \acc{\textbf{0.50}} & \acc{0.89}\\

            \midrule
            \multirow{3}{*}{\checkmark} & \multirow{3}{*}{\checkmark} & $\mathcal{P}$  & \acc{0.80} & \acc{0.61} & \acc{0.57} & \acc{0.86} \\
            & & $\mathcal{R}$  & \acc{0.88} & \acc{0.44} & \acc{0.45} & \acc{0.94} \\
            & & $\mathcal{F}1$ & \acc{\textbf{0.83}} & \acc{\textbf{0.51}} & \acc{\textbf{0.50}} & \acc{\textbf{0.90}}\\
         \bottomrule
    \end{tabular}
    \label{tab:SpecifcCategoryPerformanceBaby}
\end{table}

\begin{table}[H]

    \vspace{-0.7cm}
    
    \centering
    \caption{AutoGluon + Cleanlab performance per column in \textit{Sports}.}
    \vspace{0.1cm}
    
    \setlength\tabcolsep{7.5pt}
    \begin{tabular}{c|c|c|c|c|c|c}
        \toprule
         \textbf{Table} & \textbf{Image} & \textbf{Measure} & \textbf{Category} & \textbf{ProductType} & \textbf{Color} & \textbf{SportType} \\ \midrule

         \multirow{3}{*}{\checkmark} & \multirow{3}{*}{\xmark} & 
             $\mathcal{P}$  & \acc{0.59} & \acc{0.51} & \acc{0.39} & \acc{0.49} \\
         & & $\mathcal{R}$  & \acc{0.33} & \acc{0.56} & \acc{0.61} & \acc{0.75} \\
         & & $\mathcal{F}1$ & \acc{0.43} & \acc{0.53} & \acc{0.47} & \acc{0.59}\\
         \midrule

         \multirow{3}{*}{\xmark} & \multirow{3}{*}{\checkmark} & 
             $\mathcal{P}$  & \acc{0.77} & \acc{0.61} & \acc{0.60} & \acc{0.50} \\
         & & $\mathcal{R}$  & \acc{0.77} & \acc{0.60} & \acc{0.73} & \acc{0.71} \\
         & & $\mathcal{F}1$ & \acc{\textbf{0.77}} & \acc{\textbf{0.61}} & \acc{\textbf{0.66}} & \acc{0.59}\\
         \midrule
         
        \multirow{3}{*}{\checkmark} & \multirow{3}{*}{\checkmark} & 
             $\mathcal{P}$  & \acc{0.74} &\acc{0.62} & \acc{0.60} & \acc{0.55} \\
         & & $\mathcal{R}$  & \acc{0.74} & \acc{0.60} & \acc{0.64} & \acc{0.71} \\
         & & $\mathcal{F}1$ & \acc{0.74} & \textbf{\acc{0.61}} & \acc{0.62} & \acc{\textbf{0.62}}\\
         \bottomrule
    \end{tabular}
    \label{tab:SpecifcCategoryPerformanceSports}
\end{table}

\begin{table}[H]
    \centering
    
    \vspace{-0.7cm}
    
    \caption{AutoGluon + Cleanlab performance per column in \textit{Fashion 44K}.}
    \vspace{0.1cm}
    
    \setlength\tabcolsep{1.9pt}
    \begin{tabular}{c|c|c|c|c|c|c|c|c}
        \toprule
         \textbf{Table} & \textbf{Image} & \textbf{Measure} &  \textbf{Gender} & \textbf{Category} & \textbf{SubCategory} & \textbf{ProductType} & \textbf{Color} & \textbf{Season} \\ \midrule

        \multirow{3}{*}{\checkmark} & \multirow{3}{*}{\xmark} 
              & $\mathcal{P}$  & \acc{0.90} & \acc{1.00} & \acc{0.86} & \acc{0.74} & \acc{0.47} & \acc{0.70} \\
            & & $\mathcal{R}$  & \acc{0.08} & \acc{0.04} & \acc{0.12} & \acc{0.28} & \acc{0.21} & \acc{0.94}\\
            & & $\mathcal{F}1$ & \acc{0.15} & \acc{0.09} & \acc{0.20} & \acc{0.41} & \acc{0.29} & \acc{0.80}\\

        \midrule
        
        \multirow{3}{*}{\xmark} & \multirow{3}{*}{\checkmark} 
              & $\mathcal{P}$  & \acc{0.74} & \acc{0.96} & \acc{0.78} & \acc{0.55} & \acc{0.35}& \acc{0.44}  \\
            & & $\mathcal{R}$  & \acc{0.92} & \acc{0.99} & \acc{0.97} & \acc{0.93} & \acc{0.95}& \acc{0.77} \\
            & & $\mathcal{F}1$ & \textbf{\acc{0.82}} & \textbf{\acc{0.97}} & \textbf{\acc{0.87}} & \textbf\acc{0.70} & \textbf{\acc{0.51}}& \acc{0.57} \\

        \midrule
         
         \multirow{3}{*}{\checkmark} & \multirow{3}{*}{\checkmark} 
              & $\mathcal{P}$  & \acc{0.74} & \acc{0.96} & \acc{0.96} & \acc{0.89} & \acc{0.50} & \acc{0.75} \\
            & & $\mathcal{R}$  & \acc{0.71} & \acc{0.96} & \acc{0.72} & \acc{0.71} & \acc{0.29} & \acc{0.94}\\
            & & $\mathcal{F}1$ & \acc{0.73} & \acc{0.96} & \acc{0.82} & \textbf{\acc{0.79}} & \acc{0.37} & \textbf{\acc{0.84}}\\
         \bottomrule
    \end{tabular}
    \label{tab:SpecifcCategoryPerformanceFashionBig}
\end{table}

\newpage 
\begin{table}[H]

    \vspace{-0.2cm}

    \centering

    \caption{Error detection and repair accuracy for \textit{Fashion} columns (table + image) using Cleanlab.}
    \label{tab:SpecifcCategoryCorrectionFashion}
    \vspace{0.1cm}
    
    \setlength\tabcolsep{7.5pt}
    \begin{tabular}{c|c|c|c|c|c}
        \toprule
         \textbf{Measure} &  \textbf{Gender} & \textbf{Category} & \textbf{SubCategory} & \textbf{ProductType} & \textbf{Color} \\ \midrule
            Table       & \acc{0.01} & \acc{0.44} & \acc{0.07} & \acc{0.45} & \acc{0.06} \\
            Image       & \acc{0.25} & \textbf{\acc{0.97}} & \acc{0.66} & \acc{0.65} & \acc{0.10}\\
            Table \& Image  & \textbf{\acc{0.83}} & \acc{0.88} & \textbf{\acc{0.94}} & \textbf{\acc{0.80}} & \textbf{\acc{0.72}}\\
         \bottomrule
    \end{tabular}
    
\end{table}

\begin{table}[H]
    \centering
    \setlength\tabcolsep{10.5pt}

    \vspace{-0.2cm}

    \caption{Error detection and correction accuracy for \textit{Baby} (table + image) using Cleanlab.}
    \label{tab:SpecifcCategoryCorrectionBaby}
    \vspace{0.1cm}
    
    \begin{tabular}{c|c|c|c|c}
        \toprule
         \textbf{Measure} & \textbf{Category} & \textbf{ProductType} & \textbf{Color} & \textbf{PackageMaterial} \\ \midrule
            Table       & \acc{0.18} & \acc{0.17} & \textbf{\acc{0.21}} & \acc{0.79}  \\
            Image       & \textbf{\acc{0.79}} & \acc{0.20} & \acc{0.16} & \acc{0.18} \\
            Table \& Image  & \acc{0.54} & \textbf{\acc{0.24}} & \textbf{\acc{0.21}} & \textbf{\acc{0.81}} \\
         \bottomrule
    \end{tabular}
    
\end{table}

\begin{table}[H]
    \centering

    \vspace{-0.2cm}

    \caption{Error detection and correction accuracy  for \textit{Sports} (table + image) using Cleanlab.}
    \label{tab:SpecifcCategoryCorrectionSports}
    \vspace{0.1cm}
    
    \setlength\tabcolsep{13.8pt}
    \begin{tabular}{c|c|c|c|c}
        \toprule
         \textbf{Measure} & \textbf{Category} & \textbf{ProductType} & \textbf{Color} & \textbf{SportType} \\ \midrule
            Table       & \acc{0.15} & \acc{0.05} & \acc{0.33} & \acc{0.42}  \\
            Image       & \textbf{\acc{0.64}} & \acc{0.30} & \textbf{\acc{0.47}} & \acc{0.40} \\
            Table \& Image  & \acc{0.46} & \textbf{\acc{0.35}} & \acc{0.31} & \textbf{\acc{0.51}} \\
         \bottomrule
    \end{tabular}
   
\end{table}

\begin{table}[H]

    \vspace{-0.2cm}

    \centering
    \caption{Error detection and repair accuracy for \textit{Fashion 44K} columns (table + image) using Cleanlab.}
    \label{tab:SpecifcCategoryCorrectionFashionBig}
    \vspace{0.1cm}
    
    \setlength\tabcolsep{4.6pt}
    \begin{tabular}{c|c|c|c|c|c|c}
        \toprule
         \textbf{Measure} &  \textbf{Gender} & \textbf{Category} & \textbf{SubCategory} & \textbf{ProductType} & \textbf{Color} & \textbf{Season} \\ \midrule
            Table       & \acc{0.10} & \acc{0.04} & \acc{0.07} & \acc{0.20} & \acc{0.02} & \acc{0.67} \\
            Image       & \textbf{\acc{0.88}} & \textbf{\acc{0.98}} & \textbf{\acc{0.94}} & \textbf{\acc{0.84}} & \textbf{\acc{0.69}} & \acc{0.64}\\
            Table \& Image  & \acc{0.13} & \acc{0.73} & \acc{0.54} & \acc{0.58} & \acc{0.04} & \textbf{\acc{0.86}}\\
         \bottomrule
    \end{tabular}
    
\end{table}

\begin{table*}[h!]
    \centering
    
    \vspace{-0.4cm}

    \caption{Time elapsed for training each method \textit{Fashion} and \textit{Fashion 44K} datasets  that have similar structure and contain \num{2907} and \num{44442} items respectively. All experiments are conducted on the same machine, using two Xeon Gold 6326 CPUs at 2.9GHz, 1~TB~DDR4 memory, and an A100 80GB GPU.}
    \label{tab:FashionTimes}

    \vspace{0.1cm}
    
    \setlength\tabcolsep{22.8pt}
    \begin{tabular}{l|cc|c|c}
        \toprule
            & \textbf{Table} & \textbf{Image}& \textit{Fashion} & \textit{Fashion 44K} \\
         \textbf{Method}  & used? & used? & Time (S) & Time (S) \\ \midrule
            \method{Raha}     & \checkmark & \xmark & 12 & 85 \\
            \method{AutoGluon + DataScope}     & \checkmark & \xmark & \numprint{655} + \numprint{7} & \numprint{1782} + \numprint{4157} \\
            \method{AutoGluon + Cleanlab} & \checkmark & \xmark & \numprint{722} + \numprint{17} & \numprint{1913} + \numprint{149} \\
            \method{LLaVA (zero-shot)} & \checkmark & \xmark & \numprint{69} & \numprint{3276} \\ 
            \method{LLaVA (few-shot)} & \checkmark & \xmark & \numprint{130} &  \numprint{3724}\\ 
          \midrule
            \method{AutoGluon + DataScope}     & \xmark & \checkmark & \numprint{3036} + \numprint{321}  & \numprint{5009} + \numprint{6584} \\
            \method{AutoGluon + Cleanlab} & \xmark & \checkmark & \numprint{2749} + \numprint{353} & \numprint{5664} + \numprint{6924} \\ 
            \method{LLaVA (zero-shot)} & \xmark & \checkmark   & \numprint{1251} & \numprint{19725}\\ 
            \method{LLaVA-I. (few-shot)} & \xmark & \checkmark & \numprint{10647} & \numprint{69071} \\ 
          \midrule
           \method{AutoGluon + DataScope} & \checkmark & \checkmark & \numprint{3346} + \numprint{327} & \numprint{17289} + \numprint{5866} \\
           \method{AutoGluon + Cleanlab} & \checkmark & \checkmark & \numprint{9354} + \numprint{59}  & \numprint{19289} + \numprint{454}\\ 
           \method{LLaVA (zero-shot)} & \checkmark & \checkmark & \numprint{193} & \numprint{4366}\\ 
           \method{LLaVA-I. (few-shot)} & \checkmark & \checkmark & \numprint{497} & \numprint{8946} \\ 
           \method{LEMoN} & \checkmark & \checkmark & \numprint{489} & \numprint{5057} \\ 

         \bottomrule
    \end{tabular}
    \vspace{-0.5cm}
\end{table*}

\end{document}